%% file: iclr2026_conference.tex
\documentclass{article} % For LaTeX2e
\usepackage{iclr2026_conference,times}

\input{math_commands.tex}

\usepackage{hyperref}
\usepackage{url}
\usepackage{graphicx}
\usepackage{booktabs}   % professional horizontal rules
\usepackage{multirow}   % multirow cells
\usepackage{makecell}   % line breaks in cells
\usepackage{graphicx}
\usepackage{fontawesome}

\title{From Diffusion to Flow: Efficient Motion Generation in MotionGPT3}

\author{
Jaymin Bhan\thanks{Corresponding author. Contact: jayminbhan@gmail.com} \quad
JiHong Jeon\thanks{Equal contribution} \quad
SangYeop Jeong\footnotemark[2] \\
\\
Department of Applied Artificial Intelligence \\
Seoul National University of Science and Technology \\
\texttt{\{jayminbhan, gaebalja626, yeobi5840\}@seoultech.ac.kr} \\[6pt]
\faGithub\ \texttt{https://github.com/jayminbhan/diffusion-to-flow-MotionGPT3}
}

\iclrfinalcopy
\begin{document}

\maketitle

\begin{abstract}
Recent text-driven motion generation methods span both discrete token-based approaches and continuous-latent formulations. MotionGPT3 exemplifies the latter paradigm, combining a learned continuous motion latent space with a diffusion-based prior for text-conditioned synthesis. While rectified flow objectives have recently demonstrated favorable convergence and inference-time properties relative to diffusion in image and audio generation, it remains unclear whether these advantages transfer cleanly to the motion generation setting. 
In this work, we conduct a controlled empirical study comparing diffusion and rectified flow objectives within the MotionGPT3 framework. By holding the model architecture, training protocol, and evaluation setup fixed, we isolate the effect of the generative objective on training dynamics, final performance, and inference efficiency. Experiments on the HumanML3D dataset show that rectified flow converges in fewer training epochs, reaches strong test performance earlier, and matches or exceeds diffusion-based motion quality under identical conditions. Moreover, flow-based priors exhibit stable behavior across a wide range of inference step counts and achieve competitive quality with fewer sampling steps, yielding improved efficiency--quality trade-offs. 
Overall, our results suggest that several known benefits of rectified flow objectives do extend to continuous-latent text-to-motion generation, highlighting the importance of the training objective choice in motion priors.

\end{abstract}

\section{Introduction}

Text-driven motion generation has seen significant progress through two parallel paradigms. One prominent approach relies on discrete latent representations, where motion sequences are quantized into symbolic tokens using Vector-Quantized Variational Autoencoders (VQ-VAEs)~\citep{jiang2023motiongpthumanmotionforeign, guo2023momaskgenerativemaskedmodeling, oord2018neuraldiscreterepresentationlearning}. By treating motion as a foreign language, these methods have successfully leveraged the powerful autoregressive capabilities of large language models (LLMs). However, discretization introduces inherent limitations, including quantization artifacts and a reconstruction gap, which can suppress high-frequency details and fluid motion characteristics important for natural human movement, as noted in MotionGPT3~\citep{zhu2025motiongpt3}.
As an alternative to discrete tokenization, MotionGPT3 adopts a continuous-latent formulation by pairing a pretrained GPT-2--based text encoder with a continuous motion latent space and a text-conditioned diffusion prior for generation (see Figure~\ref{fig:architecture}). This design avoids explicit motion tokenization while retaining the semantic expressiveness of pretrained language representations, offering a compelling pathway toward continuous text-to-motion synthesis.

While diffusion-based priors have demonstrated strong performance in this setting, recent work in image and audio generation has shown that rectified flow objectives can exhibit favorable optimization and inference-time properties relative to diffusion\citep{esser2024scaling, prenger2018waveglowflowbasedgenerativenetwork}. These include earlier convergence during training and improved efficiency--quality trade-offs at inference. Whether such objective-level advantages transfer cleanly to continuous-latent motion generation, however, remains an open question. MotionGPT3 provides a natural testbed for investigating this question, as it isolates the motion prior within an otherwise fixed continuous-latent architecture.

In this work, we conduct a controlled empirical study comparing diffusion and rectified flow objectives within the MotionGPT3 framework. We replace the diffusion objective with a rectified flow objective~\citep{liu2023flow} while holding the model architecture, dataset, and evaluation protocol fixed, allowing us to isolate the effect of the generative objective on training dynamics, final performance, and inference-time behavior. Unlike diffusion, which learns to reverse a stochastic noising process, rectified flow models a direct ODE trajectory between noise and data, providing a conceptually simpler optimization path.

\newcommand{\snowflake}{\raisebox{-0.2ex}{\includegraphics[height=1em]{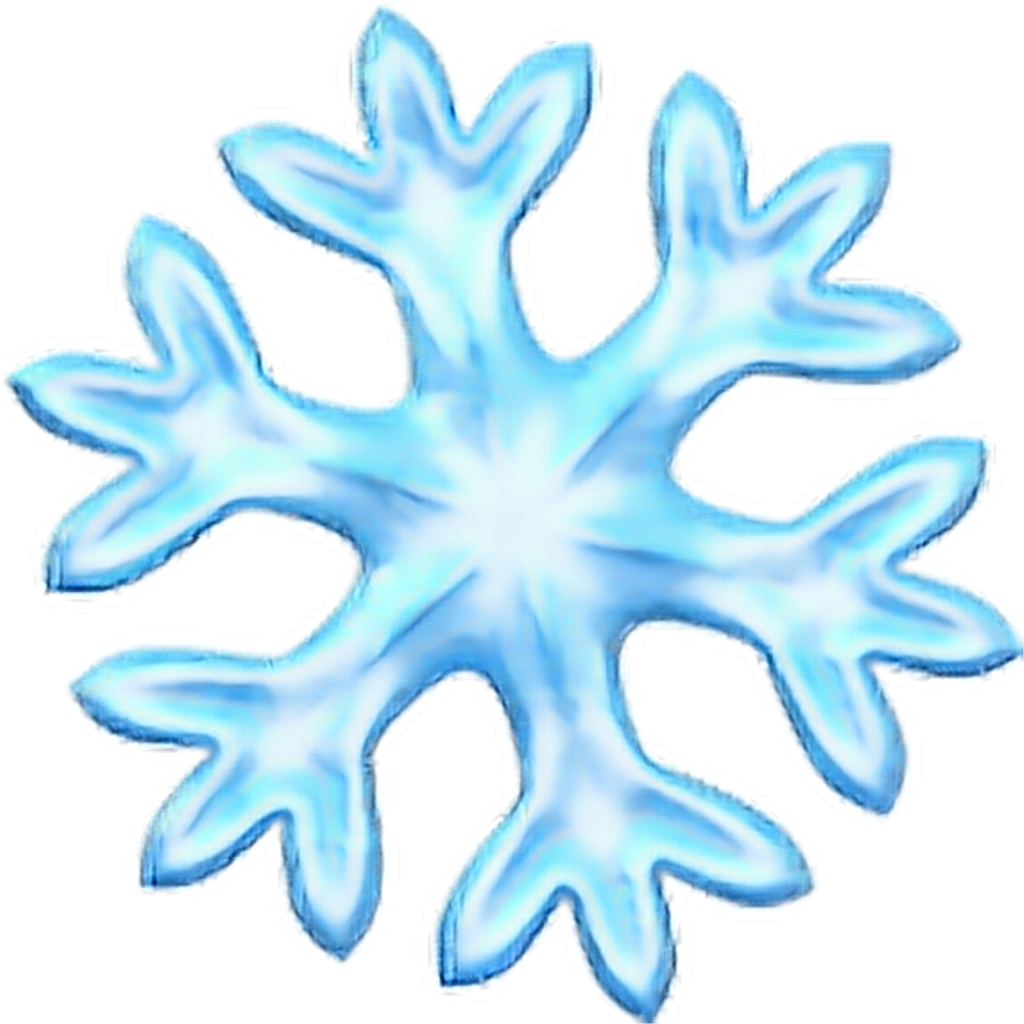}}}
\newcommand{\fire}{\raisebox{-0.2ex}{\includegraphics[height=1em]{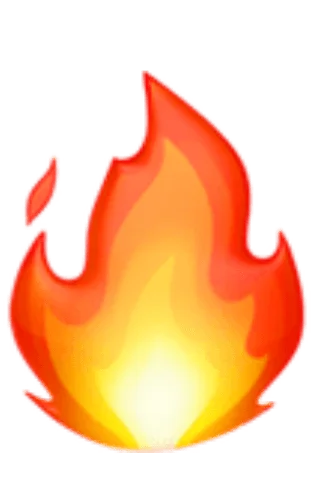}}}

\begin{figure}[t]
  \centering
\includegraphics[width=0.70\linewidth]{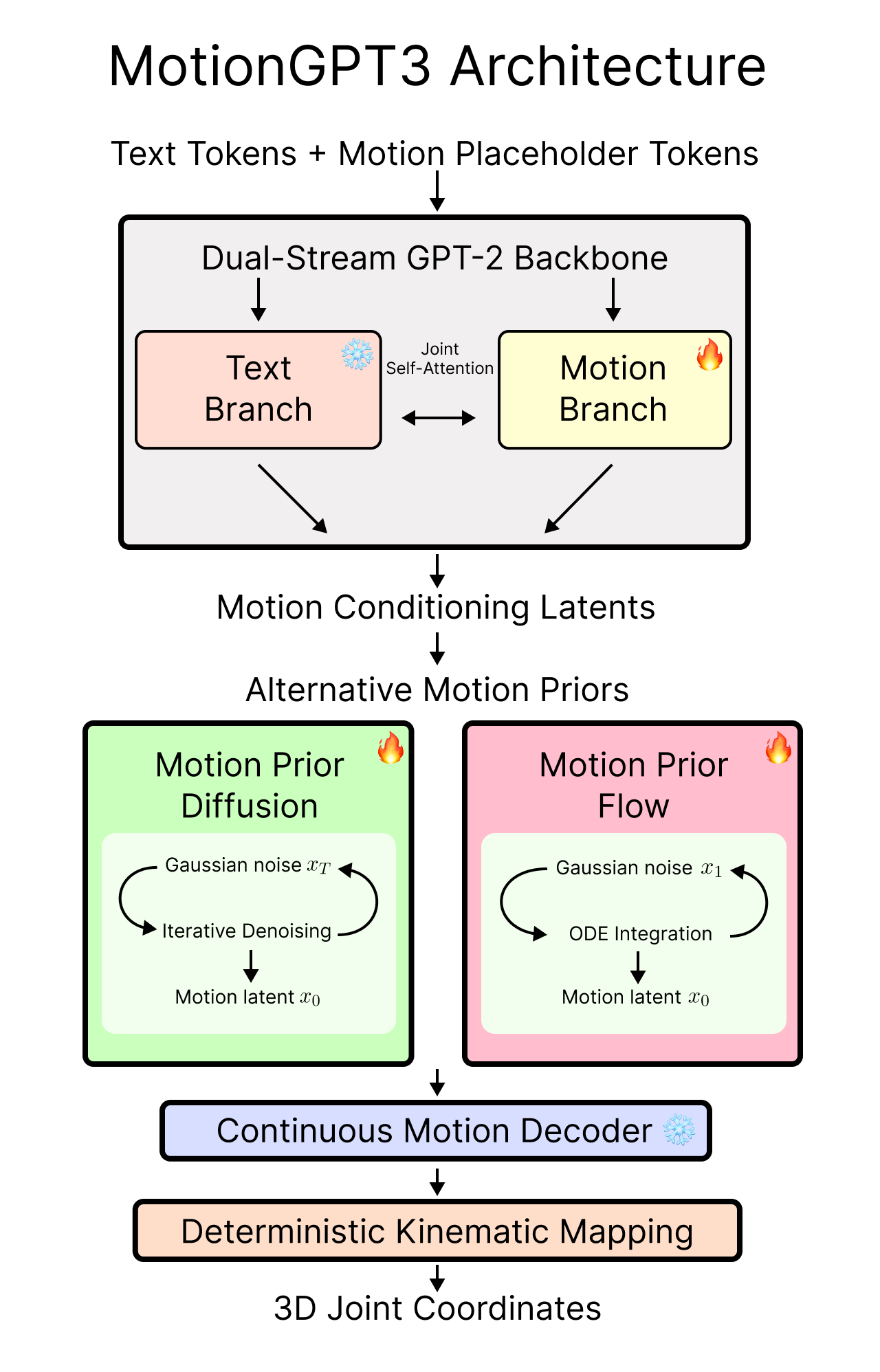}%
\caption{Overview of the MotionGPT-3 architecture with alternative motion priors. 
The dual-stream GPT-2 backbone produces text-conditioned motion latents, which are fed into either a diffusion-based or flow-based motion prior.
The \snowflake{} and \fire{} symbols indicate frozen and trainable modules, respectively.}
  \label{fig:architecture}
\end{figure}

Our experiments on the HumanML3D dataset suggest that several properties commonly associated with rectified flow in other generative domains also emerge in the motion generation setting:
\begin{itemize}
    \item \textbf{Earlier Convergence:} Rectified flow reaches strong test performance in fewer training epochs under matched training conditions.
    \item \textbf{Comparable or Improved Final Performance:} Flow-based priors match or exceed diffusion-based motion quality in terms of R-Precision and FID.
    \item \textbf{Inference-Time Robustness:} Generation quality remains stable across a wide range of inference step counts, with reduced sensitivity to step selection.
    \item \textbf{Improved Efficiency--Quality Trade-offs:} Competitive generation quality is achieved with fewer sampling steps, enabling lower inference latency.
\end{itemize}

Taken together, these findings highlight the importance of the generative training objective in continuous-latent text-to-motion models. Within the MotionGPT3 framework, our results indicate that rectified flow constitutes a practical alternative to diffusion-based priors, particularly in settings where training efficiency and inference-time latency are key considerations.

    \section{Related Work}
\label{sec:related}

\paragraph{Text-to-Motion Dataset and Evaluation Protocol}
HumanML3D ~\citep{Guo_2022_CVPR} is the current standard dataset for text-driven motion generation, providing a large-scale collection of paired motion–language data. Along with the dataset, the authors introduced a standardized evaluation framework utilizing a learned joint text–motion embedding space. This framework enables the calculation of retrieval-based metrics—such as R-Precision (R@K) and Matching Score—and has become the primary evaluation for assessing the performance and cross-modal alignment of motion generation models.

\paragraph{Motion Generation Paradigms.}
Discrete motion generation methods rely on vector-quantized representations combined
with autoregressive sequence models~\citep{zhang2023t2mgptgeneratinghumanmotion, guo2023momaskgenerativemaskedmodeling, jiang2023motiongpthumanmotionforeign}.
These approaches benefit from language-model-style generation but may be constrained by
quantization artifacts. In contrast, continuous-latent approaches model motion directly in a
continuous space, avoiding explicit tokenization and enabling finer-grained temporal
representations~\citep{tevet2023mdm, chen2023mld, zhang2022motiondiffuse}. 

\paragraph{Diffusion and Flow-Based Objectives.}
Diffusion models have demonstrated strong performance for conditional generative modeling across
multiple domains, including motion generation~\citep{ho2020denoising, tevet2023mdm, zhang2022motiondiffuse}.
However, diffusion-based sampling relies on iterative stochastic denoising, which can incur
significant inference-time cost.
Rectified flow matching~\citep{liu2023flow} has been proposed as an alternative objective that learns a
deterministic vector field transporting noise to data, and has been shown in image generation to
enable efficient sampling with fewer inference steps~\citep{esser2024scaling}.
Although flow-based objectives have been explored in motion generation, prior work typically introduces simultaneous architectural or training changes, making it difficult to isolate the effect of the generative objective itself~\citep{hu2023motionflowmatching}.

\section{Method}
\label{sec:method}
To evaluate the effect of the generative training objective in isolation, we adopt the MotionGPT3 architecture~\citep{zhu2025motiongpt3} as our experimental baseline. All architectural components, training hyperparameters, and data processing steps are kept identical to the original MotionGPT3 setup. The only modification is the choice of motion prior objective: we replace the diffusion-based prior with a rectified flow formulation, both applied to the same text-conditioned motion latents. This design enables a controlled comparison between stochastic denoising and deterministic probability flow objectives with respect to training dynamics, motion quality, and inference efficiency.

\subsection{Architectural Overview}
As illustrated in Figure~\ref{fig:architecture}, MotionGPT3 treats human motion as a trajectory within a continuous latent space and consists of three primary components:

\begin{enumerate}
    \item \textbf{Motion Latent Space:} A variational autoencoder (VAE) that compresses raw motion sequences into a continuous latent representation.
    \item \textbf{Text Conditioning:} A dual-stream GPT-2 backbone that encodes natural language prompts into a rich conditioning embedding, $c$.
    \item \textbf{Motion Prior Network:} The motion prior of the system, mapping Gaussian noise $\epsilon \sim \mathcal{N}(0, \mathbf{I})$ to motion latents $z$ conditioned on the text embedding $c$.
\end{enumerate}
Generated latents are decoded by the VAE to produce motion feature representations, which are subsequently converted to 3D joint coordinates through a deterministic kinematic post-processing step.

\subsection{Training Protocol}
All experiments are conducted on the HumanML3D dataset~\citep{Guo_2022_CVPR} with a batch size of 50.
Each training epoch corresponds to a full pass over the training split of HumanML3D.
For each sample, four timesteps are independently sampled per iteration,
producing four loss terms that are averaged to form the final training objective for both
diffusion-based and flow-based models.
We train the motion prior using the AdamW optimizer with a learning rate of $2 \times 10^{-4}$,
$\beta_1 = 0.9$, and $\beta_2 = 0.99$.
Under identical training configurations, each of the diffusion and rectified flow models requires
approximately 13 hours of training on a single NVIDIA RTX~5090 GPU.
The original MotionGPT3 training pipeline consists of three stages~\citep{zhu2025motiongpt3}. In this
work, we focus exclusively on Stage~1 training, as the primary objectives of Stage~2 and Stage~3
are aimed at improving motion understanding rather than motion generation. The original MotionGPT3
study reports only marginal gains in motion generation quality from these later stages, and their
associated objectives do not directly involve the motion prior. Accordingly, all results reported
in this paper are obtained from models trained solely under the Stage~1 setting.

\subsection{Diffusion-Based Motion Prior (Baseline)}
The motion prior is trained using a denoising diffusion probabilistic model (DDPM) objective,
which learns to predict Gaussian noise added to continuous motion latents.
Following standard DDPM practice, a scaled linear noise schedule with a fixed number of diffusion steps
is employed. During training, a timestep $t$ is sampled from the noise schedule and Gaussian noise
$\epsilon \sim \mathcal{N}(0, I)$ is added to a clean motion latent $x$ to obtain a noisy latent $x_t$.
The diffusion model predicts the noise component $\epsilon$ conditioned on the noisy latent $x_t$,
the timestep $t$, and the text conditioning representation $c$ produced by the GPT-2 encoder.
The training objective is defined as:
\begin{equation}
\mathcal{L}_{\text{diff}}
= \mathbb{E}_{x, \epsilon, t}
\left[ \| \epsilon - \epsilon_{\theta}(x_t, t, c) \|^2 \right],
\end{equation}
where $\epsilon_{\theta}$ denotes the noise prediction network.
At inference time, motion latents are generated via iterative ancestral sampling following
the DDPM reverse process, progressively denoising from pure Gaussian noise conditioned on the
input text embedding.

\subsection{Flow-Based Motion Prior}
We replace the diffusion-based training objective with a rectified flow objective,
while keeping the motion prior architecture, conditioning mechanism, and data pipeline unchanged.
Following prior work on rectified flow training, timesteps are sampled from a logit-normal
distribution, which concentrates probability mass around intermediate values (e.g., $t \approx
0.5$) and has been shown to improve training stability in large-scale generative models
\citep{esser2024scaling}. Given a clean motion latent $x_0$ and a noise sample
$x_1 \sim \mathcal{N}(0, I)$, an interpolated latent state is constructed as
\begin{equation}
x_t = (1 - t) x_1 + t x_0 .
\end{equation}

Under this formulation, the target velocity field is constant along the interpolation path and is
given by
\begin{equation}
v_{\text{true}} = x_0 - x_1 .
\end{equation}

The motion prior network takes the interpolated latent $x_t$, timestep $t$, and text conditioning
representation $c$ as input, and predicts a velocity field $v_{\theta}(x_t, t, c)$. The rectified
flow objective minimizes the squared error between the predicted and target velocities:
\begin{equation}
\mathcal{L}_{\text{flow}}
= \mathbb{E}
\left[ \left\| v_{\theta}(x_t, t, c) - v_{\text{true}} \right\|^2 \right].
\end{equation}

At inference time, motion latents are generated by integrating the learned velocity field using a
numerical ODE solver. Unlike diffusion-based sampling, which relies on iterative stochastic
denoising, rectified flow performs deterministic transport from noise to data.

\subsection{Evaluation Metrics}
We evaluate motion generation quality using standard text-to-motion metrics computed in a learned joint text--motion embedding space, following the evaluation protocol introduced with the HumanML3D dataset~\citep{Guo_2022_CVPR}. Fréchet Inception Distance (FID) measures the distributional similarity between generated and ground-truth motions in this feature space, where lower values indicate closer alignment with the ground-truth distribution. R-Precision evaluates text--motion semantic alignment by measuring how often a generated motion retrieves its corresponding text description among a set of 32 candidates, reported as top-1, top-2, and top-3 retrieval accuracy; higher values indicate stronger alignment. Matching Score computes the average Euclidean distance between correctly paired motion and text embeddings, with lower values indicating tighter semantic correspondence.
Diversity measures the overall variability of the generated motion distribution by computing the average pairwise distance between randomly sampled motion embeddings. Multimodality evaluates the diversity of multiple motions generated from the same text prompt, reflecting the model's ability to produce varied outputs. All reported results in Table~\ref{tab:performance_metric} are computed using these metrics.

\begin{table}[t]
\caption{Comparison of text-to-motion generation methods on HumanML3D.
Arrows indicate whether higher ($\uparrow$) or lower ($\downarrow$) values are better.
Real denotes statistics computed from ground-truth motion data.
Bold values indicate the best performance for each metric.}
\label{tab:performance_metric}
\centering
\small
\resizebox{\textwidth}{!}{\begin{tabular}{l l c c c c c c c}
\toprule
\textbf{Methods} &
R@1$\uparrow$ & R@2$\uparrow$ & R@3$\uparrow$ &
FID$\downarrow$ & MMDist$\downarrow$ &
Diversity$\uparrow$ & MModality$\uparrow$ \\
\midrule
Real
& 0.511 & 0.703 & 0.797 & 0.002 & 2.974 & 9.503 & -- \\
\midrule

T2M-GPT ~\citep{zhang2023t2mgptgeneratinghumanmotion}
& 0.491 & 0.680 & 0.775 & 0.116 & 3.118 & \textbf{9.761} & 1.856 \\
DiverseMotion ~\citep{lou2023diversemotiondiversehumanmotion}
& 0.515 & 0.706 & 0.802 & 0.072 & 2.941 & 9.683 & 1.869 \\
MoMask ~\citep{guo2023momaskgenerativemaskedmodeling}
& 0.521 & 0.713 & 0.807 & \textbf{0.045} & 2.958 & 9.620 & 1.241 \\
MotionGPT ~\citep{jiang2023motiongpthumanmotionforeign}
& 0.492 & 0.681 & 0.733 & 0.232 & 3.096 & 9.528 & 2.00 \\
TM2T~\citep{guo2022tm2tstochastictokenizedmodeling}
&0.424&0.618&0.729&1.501&3.467&8.589&\textbf{2.424}\\
MotionGPT3 (Reported)
& 0.533 & 0.731 & 0.826 & 0.239 & \textbf{2.797} & 9.688 & 1.560 \\
MotionGPT3 (Diffusion, ep. 142)
&0.520&0.716&0.807&0.240&2.918&9.479&2.277 \\
\textbf{MotionGPT3 (Flow. ep. 54)}
&\textbf{0.544}&\textbf{0.740}&\textbf{0.828}&0.192&2.837&9.340&2.366 \\

\midrule

\bottomrule
\end{tabular}
}
\end{table}

\begin{figure}[h]
  \centering
\includegraphics[width=0.32\linewidth]{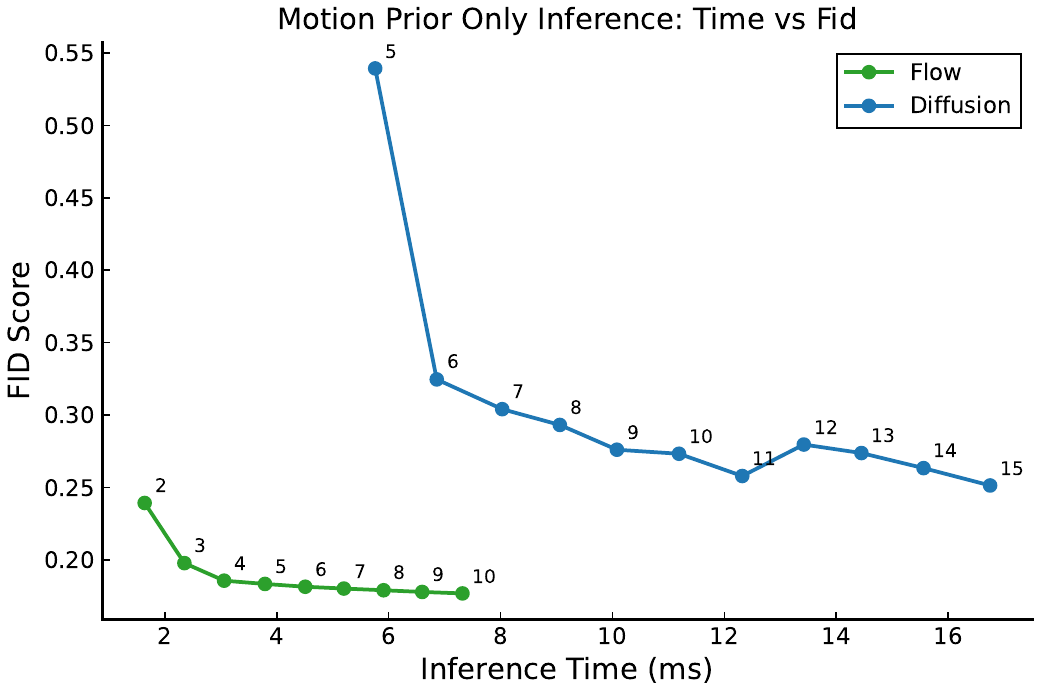}%
\includegraphics[width=0.32\linewidth]{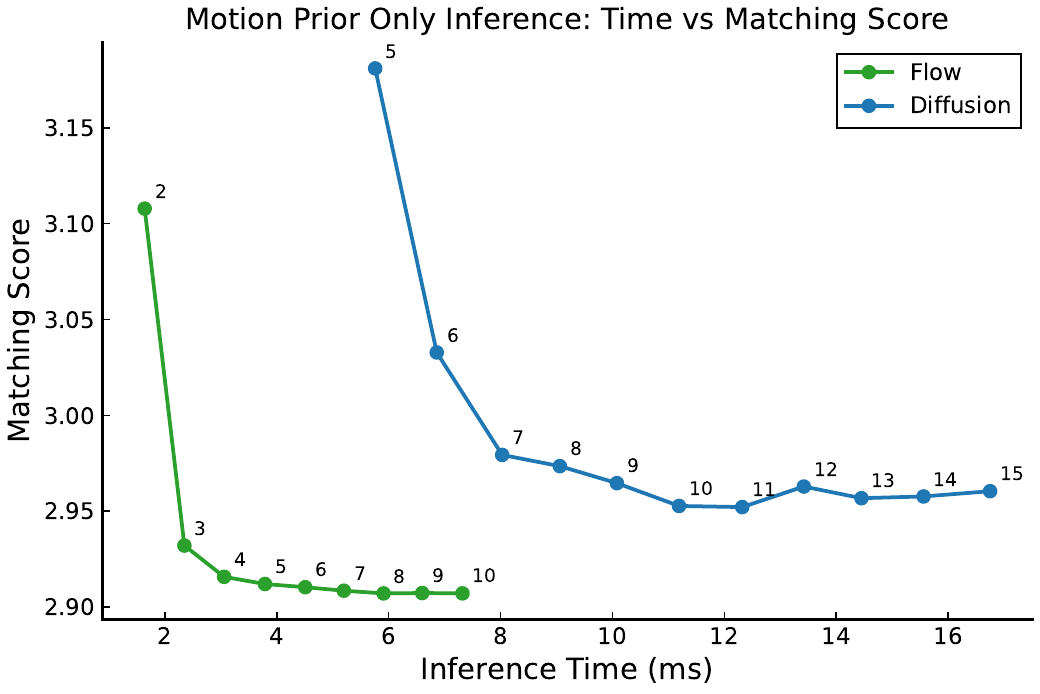}%
\includegraphics[width=0.32\linewidth]{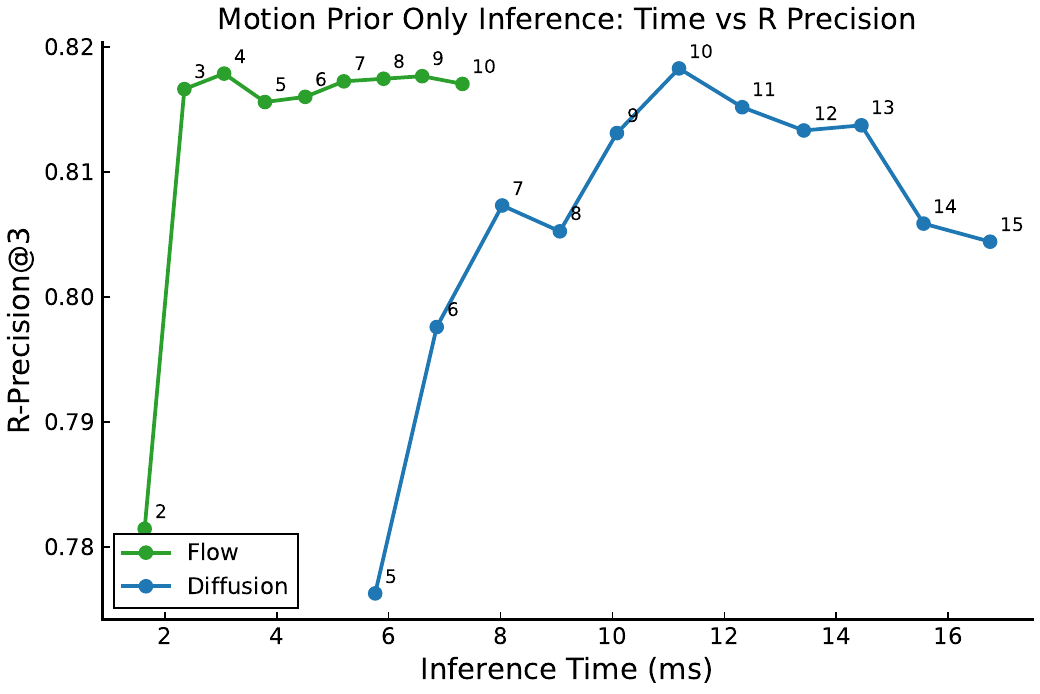}
  \caption{Motion-prior-only inference Pareto comparison between diffusion and flow variants, illustrating the trade-off between inference time and generation quality.}
  \label{fig:pareto_prior}
\end{figure}

\begin{figure}[h]
  \centering
\includegraphics[width=0.32\linewidth]{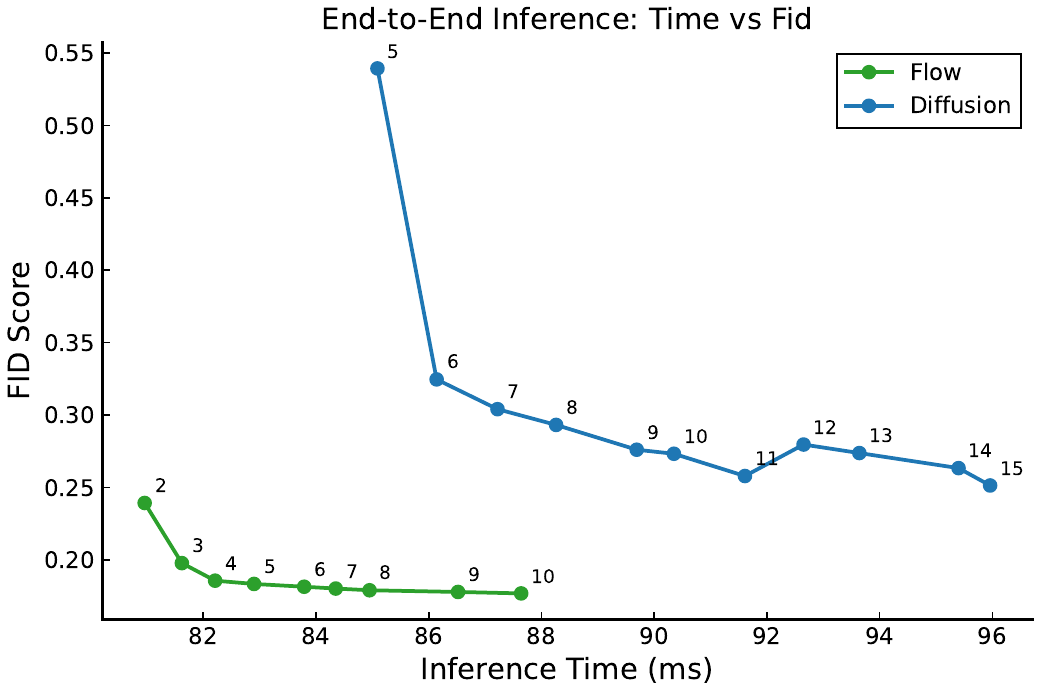}%
\includegraphics[width=0.32\linewidth]{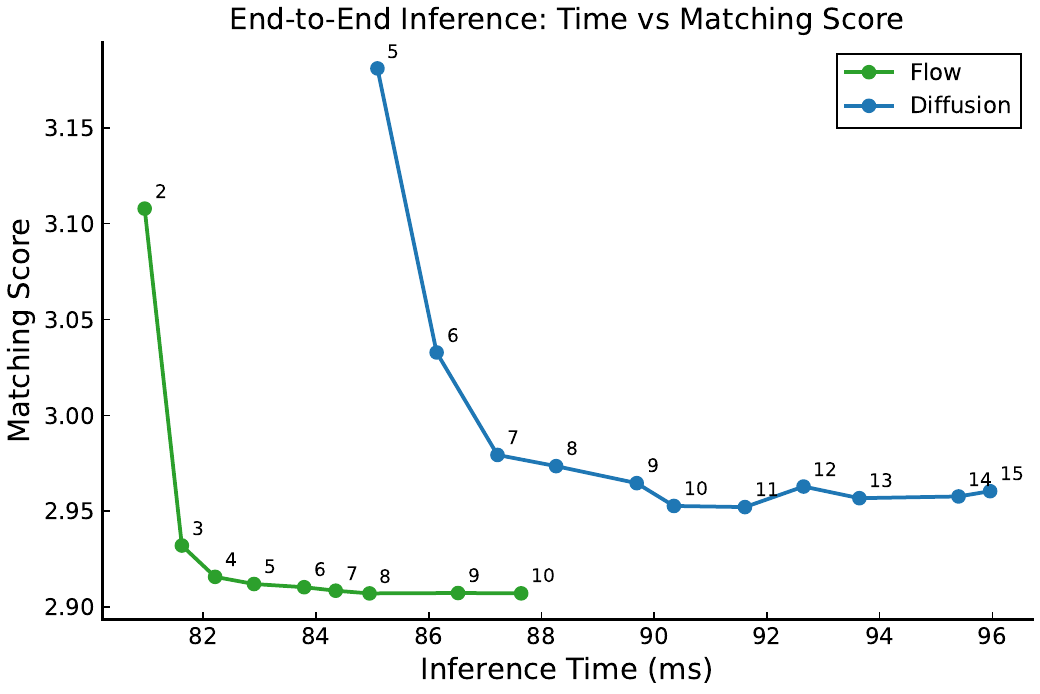}%
\includegraphics[width=0.32\linewidth]{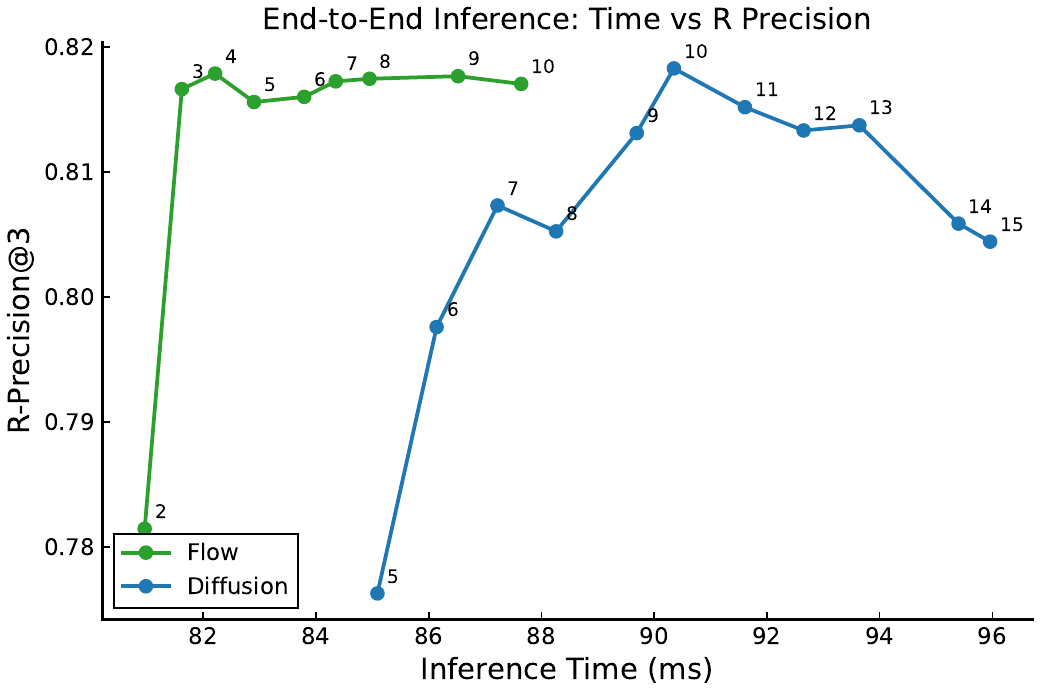}
  \caption{End to end Pareto comparison between diffusion and flow variants, illustrating the trade-off between end-to-end inference time and generation quality.}
  \label{fig:pareto_full}
\end{figure}

\newpage

\subsection{Efficiency-Quality Trade-off}
To analyze the efficiency--quality trade-off, we construct Pareto plots with inference latency on the $x$-axis and generation quality metrics (FID, Matching Score, and R-Precision@3) on the $y$-axis. Inference latency is measured using text prompts sampled from the HumanML3D training set. For each configuration, we process batches of 32 samples, performing 3 warmup iterations for CUDA memory allocation and cache stabilization, followed by 10 timed iterations. We report the average latency per batch across the timed runs. Two timing modes are measured: (1) motion-prior-only inference Figure~\ref{fig:pareto_prior}, which isolates
the latent motion generation step; and (2) end-to-end inference Figure~\ref{fig:pareto_full}, comprising
the LLM forward pass for conditioning, motion prior sampling and VAE decoding.

Inference latency is controlled by varying the number of motion prior inference steps. For the
flow-based model, motion latents are generated by integrating the learned velocity field using a
fixed-step forward Euler solver, with the number of integration steps ranging from 2 to 15. For the
diffusion baseline, we vary the number of denoising steps from 4 to 15. The diffusion model is
trained with a 1000-step noise schedule; during inference, reduced step counts are obtained by
uniformly subsampling this schedule (e.g., 100 inference steps correspond to a stride of 10).
For each step configuration, generation quality is evaluated on the HumanML3D test set using the
standard evaluation protocol described in Section~3.5.

\begin{figure}[h]
  \centering
\includegraphics[width=0.32\linewidth]{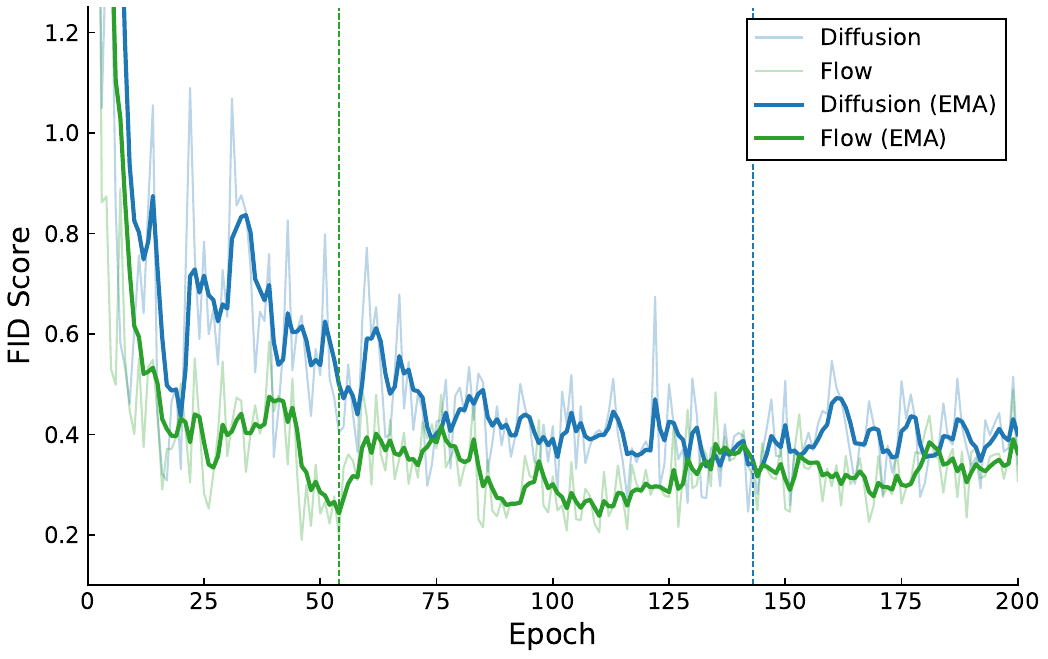}%
\includegraphics[width=0.32\linewidth]{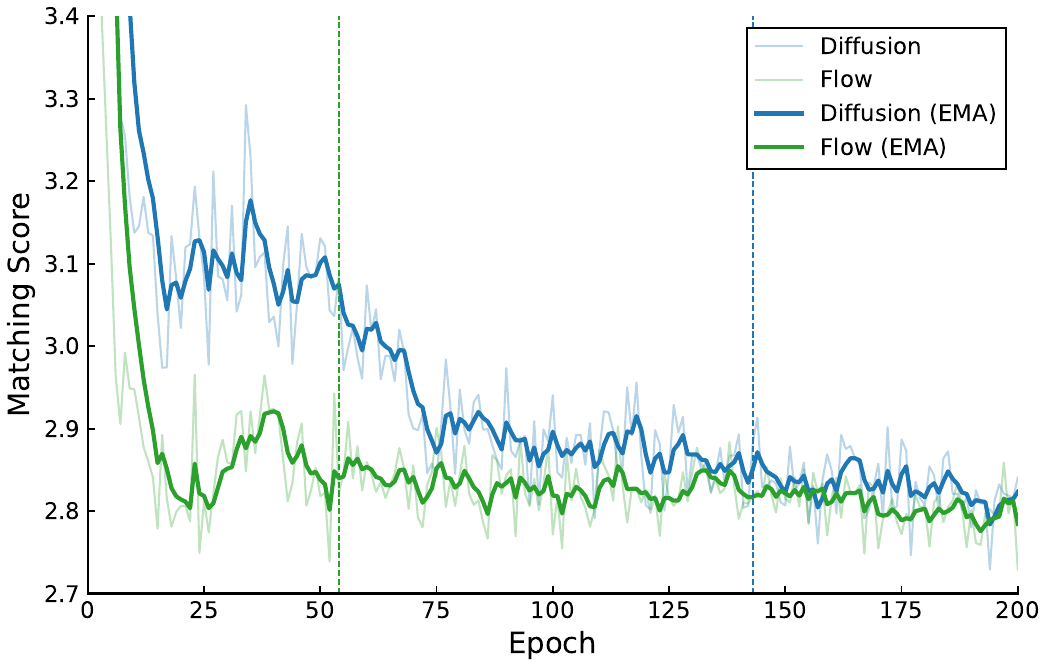}%
\includegraphics[width=0.32\linewidth]{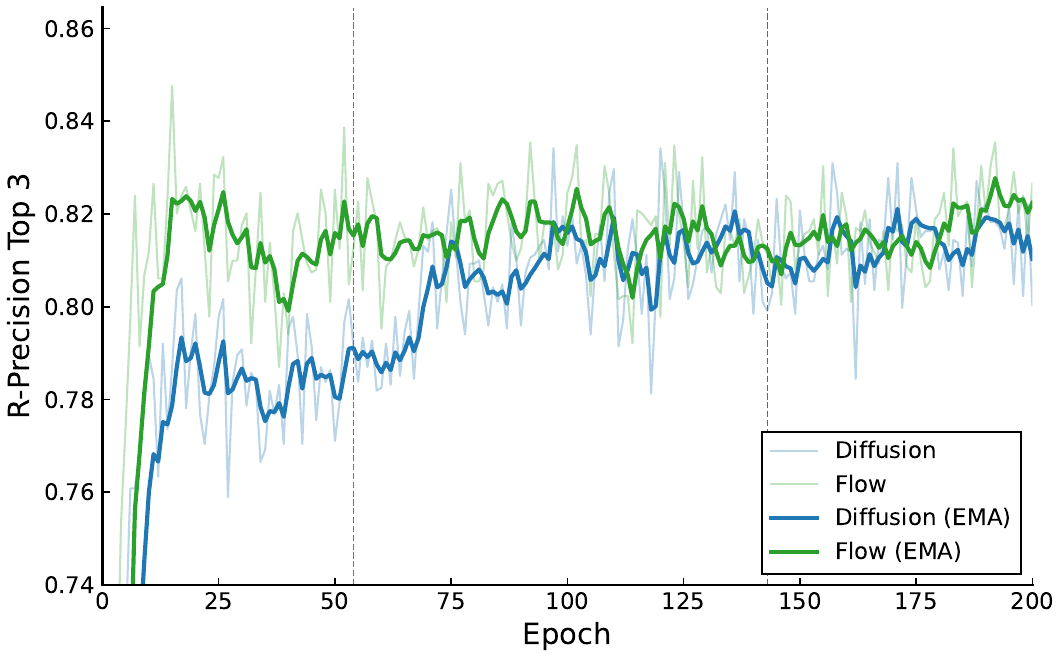}
  \caption{Validation metrics across training epochs for diffusion- and flow-based variants.
We report FID, Matching Score, and R-Precision (R@3). Faded curves indicate raw validation
measurements, while solid curves show exponential moving averages computed with a span of five
epochs to highlight overall training trends.}
  \label{fig:metrics_train}
\end{figure}

\section{Experiments}
\label{sec:experiments}
To compare training dynamics and the best validation performance achieved under different generative objectives, we evaluate both diffusion-based and flow-based motion priors on the HumanML3D validation set throughout training. Evaluation metrics are computed at every training epoch and plotted to visualize validation performance curves, enabling a direct comparison of convergence behavior and peak validation performance between the two objectives. All models are trained for 200 epochs under identical settings. We plan to release implementation details to support reproducibility.

\subsection{Validation Metric Dynamics}
During training, we analyze the evolution of validation metrics under the HumanML3D evaluation
protocol. Figure~\ref{fig:metrics_train} illustrates the validation trends for diffusion and
flow-based objectives. Across all metrics, rectified flow exhibits steeper early-stage improvement,
reaching competitive performance significantly earlier than diffusion. Validation performance for
the flow objective saturates after approximately 50 epochs, whereas diffusion continues to improve
until around 100 epochs.
We further observe that additional training stages (Stage~2 and Stage~3) yield only marginal
improvements in motion generation performance for both objectives, suggesting that the majority of
representational capacity is acquired during the initial training phase.

\subsection{Test Set Performance under Matched Conditions}
To compare the best test performance achieved by each training objective under a fixed training budget, we save model checkpoints at every epoch over 200 training epochs and evaluate each checkpoint on a held-out test set that remains fully isolated throughout training. For both diffusion- and flow-based objectives, the epoch at which peak test performance is attained differs substantially.
The flow-based model achieves its best test performance at epoch~54 (Table~\ref{tab:performance_metric}), after which additional training yields diminishing returns. In contrast, the diffusion-based model reaches its peak performance considerably later, at epoch~143. These trends are consistent with the validation dynamics observed during training. The vertical dashed lines in Figure~\ref{fig:metrics_train} indicate the epochs corresponding to peak test-set performance for each objective.
At their respective peak epochs, MotionGPT3 trained with the flow objective achieves approximately a 3\% relative improvement in R-Precision across R@1, R@2, and R@3, along with an approximately 5\% reduction in FID compared to the diffusion-based counterpart. Differences in Matching Score are comparatively small.

\subsection{Inference Latency and Speed--Quality Trade-off}
Beyond training efficiency, we evaluate inference-time latency by varying the number of sampling
steps used by the motion prior. Figure~\ref{fig:pareto_prior} reports Pareto curves that characterize
the trade-off between generation quality and inference cost attributable solely to the motion prior
across different sampling step counts.
Under identical hardware and batch-size settings, the flow-based motion prior reaches near-peak
evaluation performance with as few as four inference steps, whereas the diffusion-based prior
requires approximately eight denoising steps to attain comparable quality. In addition, even at
matched step counts, diffusion incurs higher wall-clock latency due to auxiliary post-processing
operations performed after each network invocation. In contrast, flow-based inference relies on a
lightweight numerical ODE solver with lower per-step overhead.

As a result, diffusion exhibits approximately 15--20\% longer wall-clock inference time than the
flow-based Euler solver at equivalent step counts. While the flow-based formulation
reduces the computational cost of the motion prior, end-to-end latency improvements are partially
constrained by fixed pipeline components, including text encoding and VAE decoding, which are
independent of the generative objective.
We additionally evaluated higher-order numerical solvers, such as fourth-order Runge--Kutta (RK4);
however, these did not yield meaningful improvements in evaluation quality nor reduce the effective
number of inference steps in this setting.
To account for full-pipeline effects, we further report end-to-end speed--quality trade-offs that
include text encoding, motion prior inference, and VAE decoding. As shown in Figure~\ref{fig:pareto_full},
these fixed components constitute a non-negligible fraction of total inference time, reducing the
relative latency gap between diffusion and flow. Nevertheless, the flow-based formulation
consistently achieves comparable or superior generation quality at lower end-to-end inference cost.

\section{Discussion}
\label{sec:discussion}

The experimental results indicate that, within the MotionGPT3 framework, replacing the diffusion-based motion prior with a rectified flow formulation is associated with faster convergence, improved sampling stability, and reduced inference latency under matched training and evaluation settings, while maintaining comparable overall performance. In this section, we discuss possible explanations for the observed optimization behavior, analyze the resulting efficiency--quality trade-offs, and clarify the scope and limitations of the proposed approach.

\subsection{Interpretation of Faster Convergence}
As shown in Figure~\ref{fig:metrics_train}, the motion prior trained with a rectified flow objective reaches strong validation performance in fewer training epochs than its diffusion-based counterpart across multiple metrics. In particular, flow-based training exhibits earlier stabilization of validation curves, indicating faster convergence under matched training conditions.
This behavior is consistent with the deterministic vector-field formulation of rectified flow, which models a continuous transport between noise and data distributions rather than learning to invert a stochastic noising process. By avoiding stochastic denoising trajectories during training, rectified flow reduces gradient noise associated with timestep sampling and may yield a more stable optimization landscape, allowing informative gradients to emerge earlier in training~\citep{liu2022flowstraightfastlearning}.

Taken together, these results suggest that some of the convergence advantages previously observed for rectified flow objectives in other generative settings\citep{song2023consistencymodels} can transfer to motion priors operating in continuous latent spaces.

\subsection{Efficiency--Quality Trade-offs}

In our experiments, flow-based inference attains near-peak generation quality with as few as four
Euler integration steps, whereas diffusion-based sampling requires approximately eight denoising
steps to achieve comparable performance. This difference highlights a fundamental contrast in how
the two objectives model the transformation from noise to data in continuous motion latent space.
Interestingly, we observe that the diffusion-based DDPM sampler also exhibits performance saturation
with fewer than ten denoising steps. This behavior
suggests that the underlying denoising trajectory in this domain is relatively smooth, reducing the
need for long, iterative refinement processes. In such settings, the optimal transport path between
noise and data distributions appears close to linear, favoring objectives that directly model
straight or low-curvature trajectories.

Rectified flow explicitly parameterizes this transport as a deterministic vector field and thus can
approximate the required transformation with a small number of integration steps. In contrast,
diffusion-based models learn the same mapping implicitly through stochastic denoising, requiring
multiple refinement steps to progressively remove injected noise. As a result, flow-based inference
can reach comparable solution quality with fewer steps, while diffusion must compensate through
additional iterations.

Beyond step count, flow-based inference also exhibits more stable generation quality across a range
of inference steps, whereas diffusion-based sampling is noticeably more sensitive to the number
of denoising steps. We attribute this sensitivity to the stochastic nature of diffusion sampling and
the complexity of each denoising update. Each diffusion step involves predicting noise, converting
it to a clean latent estimate, computing timestep-dependent posterior statistics, retrieving
variance coefficients from predefined schedules, and injecting additional Gaussian noise. These
operations introduce sensitivity to step discretization, which
accumulate across inference iterations. 
In contrast, flow-based inference consists of a single network evaluation followed by a simple,
deterministic update at each step. This structural simplicity results in lower per-step latency and
more predictable behavior, enabling stable performance across a wider range of step counts. Overall, these observations indicate that flow-based objectives shift the quality--efficiency Pareto
frontier in continuous-latent motion generation, achieving comparable or improved generation
quality with fewer inference steps.

\subsection{Limitations and Future Work}
Our study focuses on a controlled comparison of generative objectives within a single continuous-latent framework, MotionGPT3. While this design choice enables isolation of the training objective, it also limits the generality of our conclusions. In particular, it remains an open question whether the observed convergence and efficiency benefits of rectified flow transfer to alternative text-to-motion architectures, substantially larger motion priors, or different conditioning backbones.
Additionally, MotionGPT3 operates on a relatively compact motion latent space, which constrains the absolute latency reduction achievable through sampling step reduction alone. End-to-end inference time is further bounded by fixed pipeline components, including text encoding and VAE decoding, which are unaffected by the choice of generative objective. As a result, efficiency gains at the motion-prior level may be partially masked when measured at the full-pipeline level.

Despite these limitations, the improved sampling robustness and reduced sensitivity to inference step counts offered by flow-based priors expand the design space for latency-sensitive motion generation systems, such as interactive applications and real-time embodied agents. While it remains unclear whether the observed benefits persist uniformly as model capacity increases, scaling the motion prior represents a promising direction for future investigation. In particular, larger or more expressive motion priors may allow efficiency gains at the objective level to translate more directly into end-to-end improvements.

Finally, current standardized text-to-motion evaluations predominantly rely on explicitly motion-descriptive prompts. Extending evaluation protocols to include more implicit, narrative, or goal-oriented language—such as instructions derived from visual media or natural human interactions—would better reflect real-world deployment scenarios and place greater emphasis on efficient inference. Developing datasets and evaluation metrics aligned with such settings represents an important direction for future research.

\section{Conclusion}
In this work, we investigated how the choice of generative training objective influences optimization
behavior, inference efficiency, and motion generation quality within a continuous-latent text-to-motion framework. 
Our results show that rectified flow exhibits favorable optimization and inference properties in
continuous-latent motion generation, including faster convergence during training, stable
performance across a range of inference step counts, and competitive or improved generation
quality under matched conditions. In particular, flow-based motion priors achieve comparable
quality with fewer inference steps than diffusion-based counterparts, yielding improved
efficiency--quality trade-offs.

Taken together, these findings suggest that several benefits previously observed for flow-based
generative modeling in image and video domains transfer to text-driven motion generation when
operating in continuous latent spaces. Within the MotionGPT3 framework, rectified flow constitutes
a practical and computationally efficient alternative to diffusion-based priors, particularly for
latency-sensitive applications. We hope this study encourages further investigation of
objective-level design choices in motion generation models and their interaction with model
capacity, architecture, and downstream deployment constraints.

\section*{Acknowledgements}
In preparing this paper, we used a large language model to assist with code development and to refine the manuscript's writing. All technical content, experiments, and conclusions were developed and verified by the authors.

\bibliography{iclr2026_conference}
\bibliographystyle{iclr2026_conference}

%\appendix
%\section{Appendix}
%You may include other additional sections here.

\end{document}

%% file: math_commands.tex
\usepackage{amsmath,amsfonts,bm}

\def\eqref#1{equation~\ref{#1}}
\def\1{\bm{1}}

\DeclareMathAlphabet{\mathsfit}{\encodingdefault}{\sfdefault}{m}{sl}
\SetMathAlphabet{\mathsfit}{bold}{\encodingdefault}{\sfdefault}{bx}{n}